# MSCrackMamba: Leveraging Vision Mamba for Crack Detection in Fused Multispectral Imagery

Qinfeng Zhu
Department of Civil Engineering
Xi'an Jiaotong-Liverpool University
Suzhou, China
Qinfeng.Zhu21@student.xjtlu.edu.cn

Yuan Fang
Department of Civil Engineering
Xi'an Jiaotong-Liverpool University
Suzhou, China
Yuan.Fang16@student.xjtlu.edu.cn

Lei Fan*
Department of Civil Engineering
Xi'an Jiaotong-Liverpool University
Suzhou, China
Lei.Fan@xjtlu.edu.cn
*Corresponding Author

*Abstract*—Crack detection is a critical task in structural health monitoring, aimed at assessing the structural integrity of bridges, buildings, and roads to prevent potential failures. Vision-based crack detection has become the mainstream approach due to its ease of implementation and effectiveness. Fusing infrared (IR) channels with red, green and blue (RGB) channels can enhance feature representation and thus improve crack detection. However, IR and RGB channels often differ in resolution. To align them, higher-resolution RGB images typically need to be downsampled to match the IR image resolution, which leads to the loss of fine details. Moreover, crack detection performance is restricted by the limited receptive fields and high computational complexity of traditional image segmentation networks. Inspired by the recently proposed Mamba neural architecture, this study introduces a two-stage paradigm called MSCrackMamba, which leverages Vision Mamba along with a super-resolution network to address these challenges. Specifically, to align IR and RGB channels, we first apply super-resolution to IR channels to match the resolution of RGB channels for data fusion. Vision Mamba is then adopted as the backbone network, while UperNet is employed as the decoder for crack detection. Our approach is validated on the large-scale Crack Detection dataset Crack900, demonstrating an improvement of 3.55% in mIoU compared to the best-performing baseline methods.

I. INTRODUCTION

Bridges, roads, and buildings are critical infrastructures that require regular structural health monitoring, and crack detection plays a vital role in ensuring their safety [1]. Over time, factors such as weather, seismic activity, and general wear can degrade the integrity of these structures, leading to crack formation and posing significant safety risks. Early and accurate detection of these cracks allows for timely maintenance, preventing further damage, reducing repair costs, and ensuring the safety of the infrastructure. Traditional crack detection methods typically rely on manual inspections, which are time-consuming and inefficient. With the rapid advancements in deep learning technologies, computer vision has found widespread applications across various industries. Consequently, automated crack detection methods based on computer vision have attracted substantial attention from researchers. By analyzing data collected from sensors, deep learning-based methods can provide a more efficient and accurate assessment of the structural condition [2].

Currently, adopting deep learning networks to analyze structural information collected by visual sensors has become the most mainstream approach for crack detection due to its efficiency and ease of implementation [2]. By processing data collected from imaging devices such as cameras, deep learning methods can effectively detect and assess cracks. To enhance the crack detection capabilities of deep learning models, the fusion of infrared (IR) channels and red, green and blue (RGB) channels has demonstrated significant potential in recent years [3]. Studies have shown that the thermal information provided by IR channels can effectively improve the accuracy of crack detection [4].

Semantic segmentation is an essential approach for addressing crack detection problems. It assigns a unique label to each pixel in an image, enabling pixel-level understanding and facilitating the identification and localization of important features such as cracks within the image [5]. Therefore, semantic segmentation allows for precise localization of cracks and provides clear delineation of their patterns.

Since the introduction of convolutional neural networks (CNNs), semantic segmentation has witnessed significant advancements. Networks such as U-Net [6], DeepLab [7], and fully convolutional networks (FCNs) [8] have achieved impressive accuracy across various image segmentation tasks. These networks are designed to effectively integrate both shallow and deep features of an image, enabling them to capture fine-grained details as well as high-level semantic information. In recent years, the introduction of vision transformer (ViT) [9] has further improved segmentation performance [10]. ViT transforms the image into a sequence of patches, applying multi-head attention to each patch, which endows the network with a global receptive field, leading to enhanced contextual understanding of the image.

Despite the significant advancements in semantic segmentation networks, crack detection tasks still face numerous challenges. CNN-based methods often struggle in dealing with cracks of varying scales. This is primarily due to the limited receptive field of CNNs [11], which makes it challenging for the network to capture both detailed and global information in images with complex morphology and significant scale variation [12]. Although ViT can enhance the network's global perception capabilities, its quadratic complexity in calculating attention between patches leads to a significant increase in computational resource requirements when processing high-resolution images, resulting in limited efficiency during training and inference [9].

In addition, in the context of crack detection using fused multispectral images, a common challenge is the resolution difference between RGB and IR images. Due to the characteristics of the sensors, IR images typically have lower resolution compared to RGB images [4]. To enable semantic segmentation networks to properly process multispectral data, a common approach is to downsample RGB images to match the resolution of IR images, allowing for the fusion of both image modals into a multi-channel input, as shown in Fig. 1. However, this approach results in a loss of detail in RGB images, which negatively affects the accuracy of crack detection.

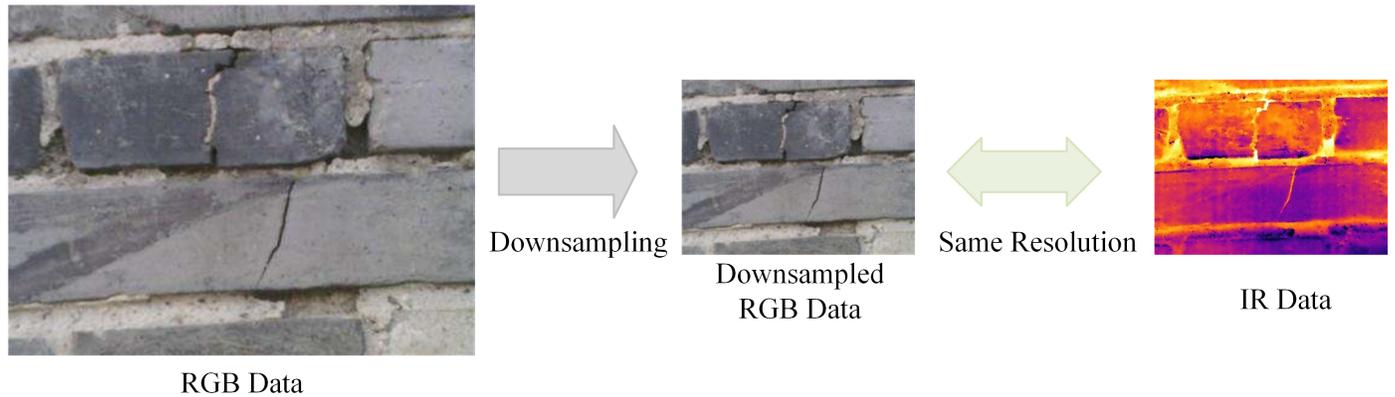

Fig. 1. Conventional method where RGB images are downsampled to align with the resolution of IR images.

To address these challenges, this paper proposes MSCrackMamba, a two-stage crack detection architecture that combines a super-resolution method with the recently introduced Mamba architecture [13]. In the first stage, a self-supervised super-resolution method is used to upsample low-resolution IR images, aligning the resolution of IR and RGB images. The RGB and IR images are then concatenated to form a six-channel input. In the second stage, the recently proposed Vision Mamba [14] is selected as the backbone network, with UperNet [15] serving as the decoder, to train on the six-channel multispectral data. Vision Mamba provides a global receptive field with linear complexity, making it well-suited for crack detection from multispectral images [16]. The main contributions of this work are as follows:

1. We propose a two-stage MSCrackMamba, a novel paradigm designed for crack detection tasks using RGB and IR images.

2. To the best of our knowledge, this is the first application of Vision Mamba to crack detection in fused multispectral images.

The remainder of this paper is organized as follows: Section II provides a detailed description of MSCrackMamba; Section III presents our experiments and results; Section IV concludes the paper and suggests future directions.

## II. METHODOLOGY

### A. Architecture Overview

MSCrackMamba is a two-stage framework designed for crack detection, with its overall architecture illustrated in Fig. 2. The primary objective of the first stage is to align the resolution of RGB and IR channels without losing the fine details of RGB images. The second stage aims to use Vision Mamba for semantic segmentation of the multi-channel multispectral data.

### B. Stage 1: Resolution Alignment

As shown in Stage 1 of Fig. 2, we propose a super-resolution approach to super-resolve IR channels, aligning their resolution with that of the RGB channels. To achieve this, we select the state-of-the-art (SOTA) super-resolution network, Fusion-Net [17]. This network is based on detail injection and utilizes a deep convolutional neural network to improve the quality of the fusion by using the difference between the ground truth and the images to be upsampled as input, thus preserving details effectively.

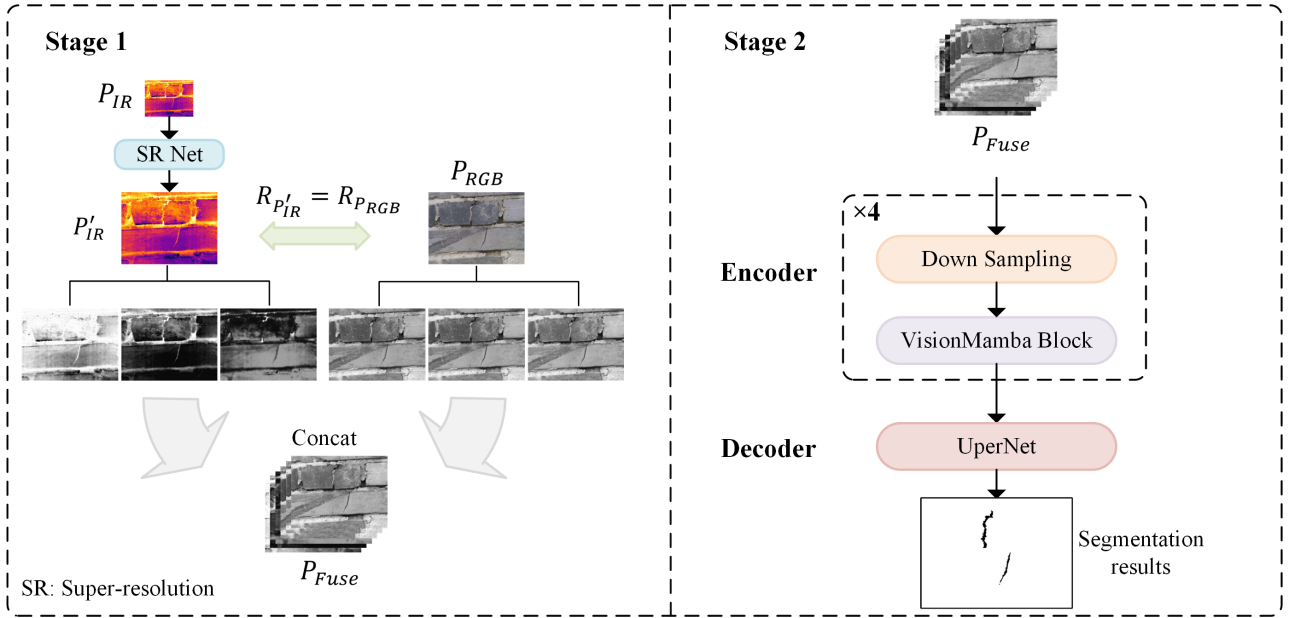

Fig. 2. Overall architecture of the MSCrackMamba framework.

We adopt a self-supervised training paradigm to super-resolve IR images [18]. We define the higher-resolution RGB data collected by the sensor as $P_{RGB}$ and the lower-resolution three-channel IR data as $P_{IR}$. Specifically, we first downsample $P_{IR}$ to obtain the downsampled $p_{IR}$. We then apply a super-resolution network to super-resolve $p_{IR}$, using $P_{IR}$ as the ground truth, thereby obtaining a trained super-resolution model $M$. Subsequently, we use $M$ to further super-resolve $P_{IR}$, resulting in $P'_{IR}$ with resolution $R_{P'_{IR}}$ matching the resolution $R_{P_{RGB}}$ of $P_{RGB}$.

After obtaining $P'_{IR}$ with the resolution aligned with $P_{RGB}$, we concatenate the three-channel $P_{RGB}$ with the three-channel $P'_{IR}$ to form the six-channel $P_{Fuse}$, which serves as the input for the subsequent semantic segmentation network.

*C. Stage 2: Semantic Segmentation*

Mamba is initially designed for large language models [13], capable of performing full-context understanding with linear complexity. It is built upon a State Space Model (SSM), which offers a robust framework for capturing dependencies in sequential data. Unlike traditional recurrent networks, which sequentially update hidden states and are prone to forgetting earlier information, SSMs maintain a continuous latent state evolution, allowing predictions to integrate information from the entire sequence. The general form of SSMs can be expressed as:

$$h'(t) = Ah(t) + Bx(t) \tag{1}$$

$$y(t) = Ch(t) + Dx(t) \tag{2}$$

where $h(t)$ represents the latent state, $x(t)$ the input, and $A, B, C, D$ define the dynamics. To adapt SSMs for discrete input, Mamba leverages the Structured State Space for Sequences (S4), employing zero-order hold discretization:

$$h_k = \bar{A}h_{k-1} + \bar{B}x_k \tag{3}$$

$$y_k = \bar{C}h_k + \bar{D}x_k \tag{4}$$

where $\bar{A} = e^{\Delta t A}$, $\bar{B} = (\bar{A} - I)A^{-1}B$, with $\Delta t$ as the sampling interval. This discretization enables Mamba to process sequential data efficiently. Additionally, Mamba incorporates a gating mechanism that adaptively controls the propagation or suppression of specific inputs, enabling the model to focus on salient features while minimizing computational overhead.

Building on the concept of ViT, which serializes images into patches, Mamba has quickly been adapted for use in the visual domain [14, 19]. Its structured state space enables comparative efficient multi-directional scanning, allowing it to learn the positional relationships of patches without relying on computationally expensive multi-head attention required by ViT. This design has demonstrated strong potential in various applications, including remote sensing [12, 20].

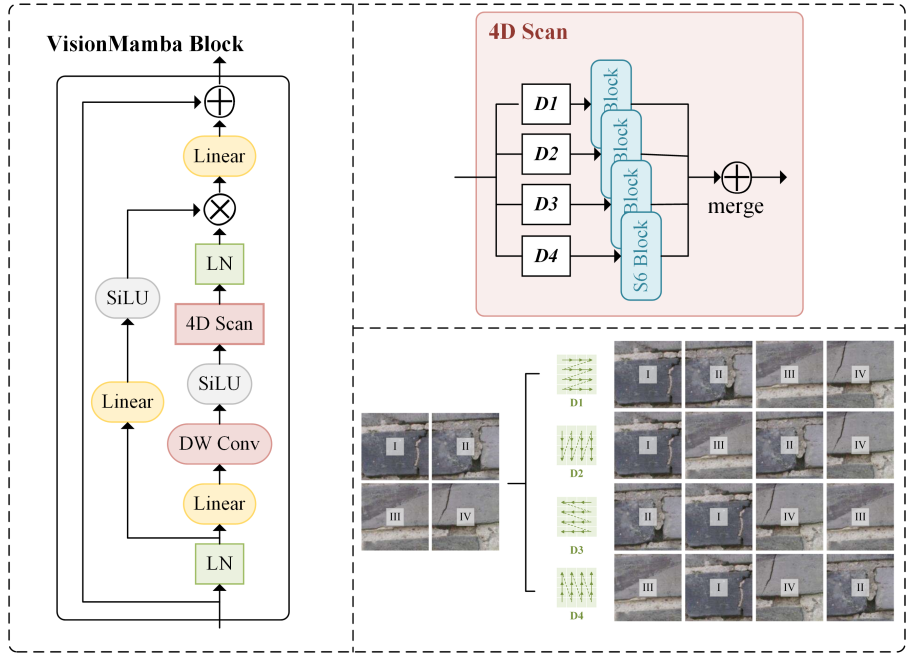

Fig. 3. The framework of the VisionMamba Block. LN：Layer Normalization. DW Conv: Deepwise Convolution.

As shown in Stage 2 of Fig. 2, we use the six-channel $P_{Fuse}$ data as input to the semantic segmentation network. We constructed the network using an encoder-decoder architecture, which is widely used in semantic segmentation tasks. Four VisionMamba Blocks are used to build the encoder, and the input data undergoes four stages of downsampling. The SOTA UperNet [15] is used to build the decoder, which reconstructs the segmentation map from multiple levels of features.

Fig. 3 illustrates the VisionMamba Block, which is a residual network with skip connections. One branch employs a linear mapping with SiLU activation, while the other branch uses depthwise convolution to extract information. These features undergo scanning in four different directions using S6 computation [13], and the outputs are then merged.

Similar to ViT, VisionMamba partitions the image into patches and flattens each patch as a sequence. However, unlike ViT, VisionMamba does not compute multi-head attention between patches. Instead, multi-directional scanning was used to allow the Mamba architecture to learn the positional information of image patches. The illustration of multi-directional scanning is shown in the lower right corner of Fig. 3, with the four directions being left to right, top to bottom, right to left, and bottom to top. In practice, we typically use a block size of 3 pixels for partitioning.

III. EXPERIMENTS

*A. Dataset*

We selected a recent large-scale multimodal crack detection dataset, Crack900 [4], to validate the MSCrackMamba architecture. The dataset contains 914 finely annotated RGB and IR images of masonry structure cracks, which were captured using a FLIR E85 IR camera at the ancient city walls in Suzhou, China. The IR sensor resolution is 384×288, while the RGB sensor resolution is 1280×960. The dataset was randomly divided into a training set (80%) and a test set (20%), resulting in 731 training images and 183 validation images.

We used the $mIoU$ (mean Intersection over Union) metric to evaluate the segmentation accuracy, which is calculated using the following formula:

$$mIoU = \frac{1}{N} \sum_{i=1}^{N} \frac{TP_i}{TP_i + FP_i + FN_i} \tag{5}$$

where $N$ represents the total number of classes, which is generally 2 in the crack detection dataset, including crack and background classes. $TP_i$ represents the number of true positive pixels for class $i$, $FP_i$ represents the number of false positive pixels for class $i$, and $FN_i$ represents the number of false negative pixels for class $i$. A higher $mIoU$ score indicates better overall segmentation performance [21].

*B. Training Settings*

In order to ensure a fair comparison, our network parameter settings were kept as consistent as possible with those used in the benchmark networks of the article introducing Crack900, as shown in TABLE I. The only difference is that we increased the patch size from 256×256 to 512×512. This adjustment was made because we super-resolved the IR channels to match the resolution of the RGB channels (1280×960), allowing the network to crop at a larger scale. Data augmentation techniques were employed to enhance the generalization ability of the network [22], including Random Flip, Random Rotate, and Random Crop. All experiments were conducted using two 4090D GPUs (24G), with a batch size of 8 per GPU.

We adopted the pretraining-finetuning approach in our experiments to ensure optimal segmentation performance. This is a common training strategy in semantic segmentation tasks. It involves first training the encoder-decoder structure on a large-scale image dataset and then fine-tuning it on the downstream segmentation task. The pretrained encoder can more effectively extract features from images, thereby improving performance in downstream tasks.

TABLE I. TRAINING SETTINGS FOR MSCRACKMAMBA FRAMEWORK

| | |
|---|---|
| Patch size | 512×512 |
| Total learning iterations | 20000 |
| Batch size | 8×2 |
| Optimizer | AdamW |
| Weight decay | 0.01 |
| Schedule | PolyLR |
| Warmup | 1500 iterations |
| Learning rate | 0.00003 |
| Loss function | Cross entropy |

*C. Segmentation Results*

All experiments were retrained to eliminate the influence of training equipment and parameters, ensuring a fair comparison. TABLE II summarizes the experimental results. The *mIoU* score achieved using the MSCrackMamba architecture was 76.96%, which represents a significant improvement of 3.55% compared to the best-performing combination of ConvNeXt-t and UperNet. This demonstrates the effectiveness of the introduced MSCrackMamba architecture.

Examples visualized segmentation results are shown in Fig. 4, where we compared MSCrackMamba with the previously best-performing combination of ConvNeXt and UperNet. It can be observed that the earlier methods exhibited noticeable false positive errors (as shown in Fig. 4(b), (c), and (d)) and false negative errors (as indicated in Fig. 4(a) and (e)). In contrast, the MSCrackMamba architecture demonstrated a significantly better capacity for capturing the shape of the cracks, although occasional minor false positive errors are still present (e.g., Fig. 4(e)).

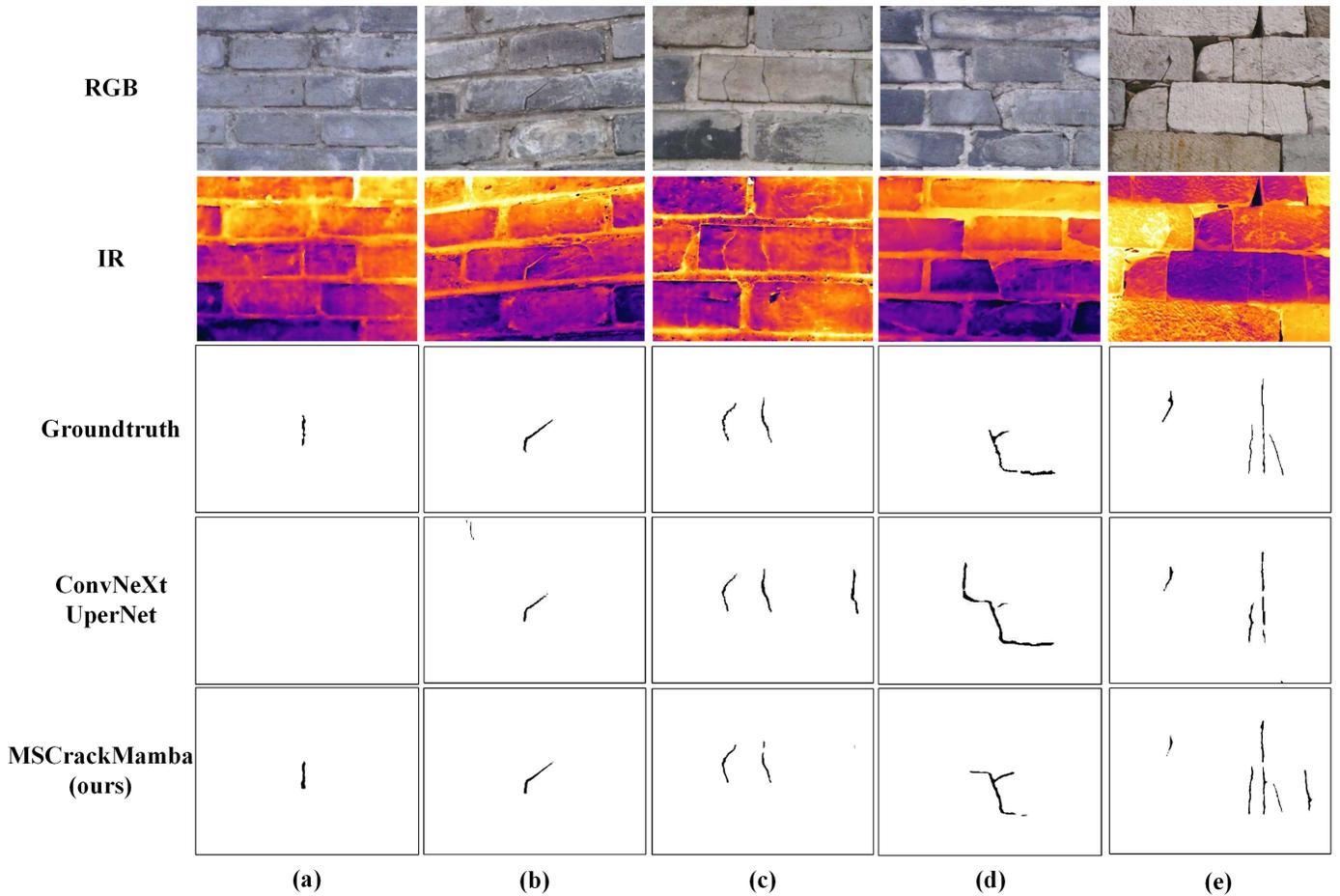

Fig. 4. Visual comparisons of segmentation outcomes between our MSCrackMamba and the best-performing combination of ConvNeXt and UperNet.

*D. Alabation Experiments*

In our ablation experiments, we aimed to verify the effectiveness of the two-stage strategy and/or pretraining. In these tests, we used VisionMamba as the backbone and UperNet as the decoder. The results, as shown in TABLE III, indicate that both the two-stage strategy and pretraining contribute to improving segmentation performance. When they were combined, an improvement of 4.42% in $mIoU$ was observed compared to the baseline without either. Due to the strong performance of the Mamba-based backbone, even without the two-stage training strategy or pretraining, the segmentation results still outperformed previous works.

TABLE II. ACCURACY OF SEMANTIC SEGMENTATION ON THE CRACK900 VALIDATION SET FROM OUR MSCRACKMAMBA FRAMEWORK AND OTHER COMPARED METHODS. THE HIGHEST SCORE IS HIGHLIGHTED IN BOLD.

| Strategy | Encoder | Decoder | mIoU /% |
|---|---|---|---|
| | ConvNeXt-t [23] | UperNet [15] | 73.41 |
| | ResNet50 [24] | DeepLabV3+ [7] | 72.02 |
| | ResNet50 [24] | FCN [8] | 70.52 |
| Original | ResNet50 [24] | PSPNet [25] | 66.91 |
| | Swin-tiny [26] | UperNet [15] | 73.15 |
| | ResNet50 [24] | UperNet [15] | 71.01 |
| | ResNet50 [24] | BiSeNet [27] | 71.50 |
| 2-stage | VisionMamba-t | UperNet [15] | **76.96** |

TABLE III. ACCURACY OF SEMANTIC SEGMENTATION ON THE CRACK900 VALIDATION SET USING MSCRACKMAMBA FRAMEWORK WITH AND WITHOUT 2-STAGE AND PRETARINING. THE HIGHEST SCORE IS HIGHLIGHTED IN BOLD.

| 2-stage | Pretrain | mIoU /% |
|---|---|---|
| ✗ | ✗ | 72.54 |
| ✗ | ✓ | 75.47 |
| ✓ | ✗ | 74.18 |
| ✓ | ✓ | **76.96** |

To further validate the effectiveness of the two-stage strategy beyond the Mamba architecture, we also conducted experiments with CNN-based and ViT-based architectures. The experimental results are summarized in TABLE IV, where a notable improvement in segmentation accuracy was observed for all models when the two-stage strategy was employed. This further confirmed the advantage of the two-stage strategy and its generality. The combination of BiSeNet and UperNet showed the most significant enhancement, with a 4% increase in $mIoU$.

TABLE IV. ACCURACY OF SEMANTIC SEGMENTATION ON THE CRACK900 VALIDATION SET WITH AND WITHOUT 2-STAGE STRATEGY. THE HIGHER SCORES ARE HIGHLIGHTED IN BOLD.

| 2-stage | Encoder | Decoder | mIoU /% |
|---|---|---|---|
| ✗ | ConvNeXt-tiny | UperNet | 75.18 |
| ✓ | ConvNeXt-tiny | UperNet | **76.6 (+1.42)** |
| ✗ | ResNet50 | DeepLabV3+ | 73.59 |
| ✓ | ResNet50 | DeepLabV3+ | **74.94 (+1.35)** |
| ✗ | ResNet50 | FCN | 69.79 |
| ✓ | ResNet50 | FCN | **71.04 (+1.25)** |
| ✗ | ResNet50 | PSPNet | 70.58 |
| ✓ | ResNet50 | PSPNet | **74.4 (+3.82)** |
| ✗ | Swin-tiny | UperNet | 74.48 |
| ✓ | Swin-tiny | UperNet | **76.31 (+1.83)** |
| ✗ | ResNet50 | UperNet | 73.79 |
| ✓ | ResNet50 | UperNet | **74.48 (+0.69)** |
| ✗ | ResNet50 | BiSeNet | 70.48 |
| ✓ | ResNet50 | BiSeNet | **74.48 (+4.00)** |

To verify the effectiveness of integrating IR images, we evaluated the performance of the VisionMamba-t and UperNet combination on several variations of input data, including: downsampled RGB images alone (denoted as $p_{RGB}$), original resolution RGB images alone ($P_{RGB}$), a combination of downsampled RGB and original resolution IR images ($p_{RGB} + P_{IR}$), and our proposed combination of original resolution RGB with super-resolved IR images ($P_{RGB} + P'_{IR}$). The segmentation performance is summarized in TABLE V. It can be observed that integrating super-resolved IR images resulted in a $mIoU$ improvement of 5.84% compared to using only the original resolution RGB channels. Although the original strategy ($p_{RGB} + P_{IR}$) did not perform as well as our proposed two-stage approach, it still demonstrated the effectiveness of incorporating IR imagery.

TABLE V. ACCURACY OF SEMANTIC SEGMENTATION ON THE CRACK900 VALIDATION SET USING VARIOUS INPUT DATA WITH THE COMBINATION OF VISIONMAMBA-T AND UPERNET. THE HIGHEST SCORE IS HIGHLIGHTED IN BOLD.

| Input | mIoU /% |
|---|---|
| $p_{RGB}$ | 71.10 |
| $P_{RGB}$ | 71.12 |
| $p_{RGB} + P_{IR}$ | 75.47 |
| $P_{RGB} + P'_{IR}$ (ours) | **76.96** |

## IV. CONCLUSION AND FUTURE WORK

This paper introduces MSCrackMamba, a two-stage framework specifically designed for crack detection in multispectral images. The first stage ensures resolution alignment between multispectral channels while preserving the fine details of RGB channels through super-resolution of IR channels. The second stage involves the implementation of VisionMamba, leading to linear complexity and more effective capture of global contextual relationships in multispectral images. Experiments have been carried out to quantitatively investigate the contribution of each stage for enhancing crack detection. By integrating super-resolution techniques with the VisionMamba network, MSCrackMamba achieved significant performance improvements on the large-scale multispectral Crack900 dataset, outperforming the best baseline (from typical CNN and ViT-based networks) by 3.55%.

The directions for future research include:

1. More accurate super-resolution methods: While the current two-stage approach achieves multichannel resolution alignment, the self-supervised super-resolution process may introduce distortion, leading to misalignment of features and thus affecting segmentation accuracy. Future work can focus on enhancing the accuracy of super-resolution to mitigate such issues.

2. Lightweight VisionMamba backbone: Although Mamba has linear computational complexity, the VisionMamba backbone still employs four directional scans, which is computationally intensive. Studies [20] suggest that reducing the number of scan directions has minimal impact on performance. Future research could explore lightweight design strategies [28], such as alternating scan directions akin to Vision LSTM [29, 30], to make the architecture more efficient.

3. End-to-end training: The current two-stage approach is operationally time-consuming. Future work could investigate end-to-end networks that allow resolution alignment and semantic segmentation to be trained simultaneously, enhancing efficiency.

4. Enhanced multispectral channel fusion: In this study, we utilized six-channel data, limiting the spectral resolution. However, in future applications, higher spectral resolutions, such as hyperspectral data, may be encountered. In such cases, the VisionMamba backbone may not be suitable. Therefore, future research should focus on optimizing the fusion of multispectral channels [31, 32] to enable the network to adaptively adjust weights according to the features of each channel, improving the model's adaptability to complex environments.

FUNDING

This work was supported by the Xi'an Jiaotong-Liverpool University Research Enhancement Fund under Grant REF-21-01-003.

DECLARATION OF COMPETING INTEREST

The authors declare that they have no known competing financial interests or personal relationships that could have appeared to influence the work reported in this paper.